\documentclass[sn-apa]{sn-jnl}

\jyear{2021}%

\theoremstyle{thmstyleone}%
\theoremstyle{thmstyletwo}%

\theoremstyle{thmstylethree}%
\usepackage{lineno}
\usepackage{natbib}
\setcitestyle{square, comma, numbers,sort&compress, super}
\usepackage{amsmath}
\usepackage{amsfonts}
\usepackage{amssymb}
\usepackage{amsbsy}
\usepackage{color,soul}

\usepackage{xspace}
\usepackage{amssymb}
\usepackage{newtxmath}
\usepackage{float}
\usepackage{url}
\usepackage{booktabs, makecell}
\usepackage{cuted}
\usepackage{hyperref}

\usepackage[printonlyused,withpage]{acronym}
\usepackage{enumitem}

\begin{document}

 \title[AOD Fusion for PM2.5 Estimation]{Data level and decision level fusion of satellite multi-sensor AOD retrievals for improving PM2.5 estimations, a study on Tehran}


\author*[]{\fnm{Ali} \sur{Mirzaei},\fnm{Hossein} \sur{Bagheri},\fnm{Mehran}\sur{Sattari}}\email{h.bagheri@cet.ui.ac.ir}



\affil[]{\orgdiv{Faculty of Civil Engineering and Transportation}, \orgname{University of Isfahan}, \orgaddress{\street{Azadi square}, \city{Isfahan}, \postcode{8174673441}, \country{Iran}}}




\abstract{\textcolor{red}{This is the pre-acceptance version, to read the final version, please go to Earth Science Informatics on Springer, URL: \url{https://link.springer.com/article/10.1007/s12145-022-00912-6}}. One of the techniques for estimating the surface particle concentration with a diameter of fewer than 2.5 micrometers (PM2.5) is using aerosol optical depth (AOD) products. Different AOD products are retrieved from various satellite sensors, like MODIS and VIIRS, by various algorithms, such as Deep Blue and Dark Target. Therefore, they don’t have the same accuracy and spatial resolution. Additionally, the weakness of algorithms in AOD retrieval reduces the spatial coverage of products, particularly in cloudy or snowy areas. Consequently, for the first time, the present study investigated the possibility of fusing AOD products from observations of MODIS and VIIRS sensors retrieved by Deep Blue and Dark Target algorithms to estimate PM2.5 more accurately. For this purpose, AOD products were fused by machine learning algorithms using different fusion strategies at two levels: the data level and the decision level. First, the performance of various machine learning algorithms for estimating PM2.5 using AOD data was evaluated. After that, the XGBoost algorithm was selected as the base model for the proposed fusion strategies. Then, AOD products were fused. The fusion results showed that the estimated PM2.5 accuracy at the data level in all three metrics, RMSE, MAE, and $R^{2}$, was improved ($R^{2}$=0.64, MAE=9.71$\frac{\mu g}{m^{3}} $ , RMSE=13.51$\frac{\mu g}{m^{3}} $). Despite the simplicity and lower computational cost of the data level fusion method, the spatial coverage did not improve considerably due to eliminating poor quality data through the fusion process. Afterward, the fusion of products at the decision level was followed in eleven scenarios. In this way, the best result was obtained by fusing Deep Blue products of MODIS and VIIRS sensors ($R^2$=0.81, MAE=7.38$ \frac{\mu g}{m^{3}} $ , RMSE =10.08$ \frac{\mu g}{m^{3}} $). Moreover, in this scenario, the spatial coverage was improved from 77\% to 84\%. In addition, the results indicated the significance of the optimal selection of AOD products for fusion to obtain highly accurate PM2.5 estimations.}




\maketitle

\section{Introduction}\label{sec.intro}
PM2.5 is one of the most dangerous and harmful air pollutants for human health \cite{RN1}, which causes respiratory diseases, heart diseases, and premature death due to its easy absorption into the bloodstream \cite{RN2}. Therefore, measuring and monitoring the concentration of these particles is essential for the health care of people at risk. Using ground air quality control stations to measure PM2.5 is a standard method of PM2.5 monitoring. These stations have high construction and maintenance costs. Also, some of them may also not report PM2.5 concentrations in some time periods due to technical issues. However, due to the insufficient ground stations for measuring PM2.5, particularly in suburbs, it is not possible to accurately estimate concentrations of these particles, especially in areas with high variabilities, as well as suburb areas. As a solution, AOD products from satellite observations can be used to estimate these pollutants. Because of extensive spatial coverage and temporal continuity, satellite observations are more cost-effective than ground measurements. This is achieved by determining the optical depth of the atmosphere induced by the concentration of aerosol particles in the atmosphere. Thus, AOD can be used as an alternative indicator in studying air quality and PM2.5 particle concentration in case of a lack of ground measurement.

Many studies have been performed to estimate the concentration of PM2.5 from AOD products \cite{RN7,RN4,RN3,RN5,RN6}. In addition to AOD products, meteorological data are used as an influential factor in calculating the concentration of PM2.5 \cite{RN9,RN7,RN3}. Parameters such as wind speed and direction, relative humidity, temperature, and air pressure in the vertical layers of the atmosphere directly affect the emission of pollutant particles \cite{RN10}. Previous studies have also illustrated a substantial relationship between meteorological data and PM2.5 concentrations \cite{RN9,RN7,RN10,RN11}. 

So far, several methods have been used to determine the relationship between PM2.5 and AOD. Some of these methods are based on machine learning, which has proven their capability in numerous studies \cite{RN7,RN17,RN18,RN19,RN23,RN21,RN4,RN3,RN22,RN15,RN20,RN13}
 \cite{RN4}
 \cite{RN13}
\cite{RN15}
\cite{RN17}
\cite{RN7}. The various algorithms used in previous studies indicate the importance of algorithm selection for modeling the PM2.5-AOD relationship. Consequently, in this investigation, the ability of different machine learning algorithms to estimate PM2.5 from AOD products and meteorological data is evaluated in the first step. 

In addition to the algorithm used in modeling the relationship between PM2.5 and AOD, the quality of input data, especially AOD data, significantly impacts the accuracy of PM2.5 concentration estimation. Various AOD retrieval algorithms have recently been developed \cite{RN24}. Different AOD retrieval algorithms do not always produce the same accuracy and spatial resolution output. They have various performances in different regions \cite{RN25}. For example, the Dark Target (DT) product performs better for areas where surface reflection is low in the visible spectrum, such as vegetated areas \cite{RN26}. On the other hand, Deep Blue (DB) achieves more accurate results on bright surfaces such as urban areas \cite{RN27}.However, there are many uncertainties in the above products. These uncertainties arise from the many assumptions and approximations adopted during the AOD retrieval process, including the cloud screening sections, the selection of the retrieval model, and the determination of surface reflectance. In addition, the spatial coverage of AOD products in cloudy or snowy areas is reduced, resulting in non-value pixels in these products, which subsequently makes the estimation of PM2.5 impossible in these areas. As a result, the product obtained from the observations of a single sensor cannot provide continuous measurement due to the lack of continuous spatial and temporal coverage or insufficient accuracy \cite{RN28}. Also, due to differences in viewing angles, differences in spectral channels, and the spatial and temporal resolution of different sensors, there are differences in their retrieved AOD products \cite{RN30}. 

Differences in the performance of various products have been shown in previous investigations. Ning Liu et al. compared the performance of MAIAC, MDT (MODIS Dark Target), and MDB (MODIS Deep Blue) products in China. The results showed that the MAIAC product was overall superior to MDB and MDT. However, the MDT product performed better in agricultural and forestry areas than the other two. The study also demonstrated that geometric parameters such as solar zenith angle, scattering angle, and relative azimuth angle significantly affected the performance of all three algorithms in AOD retrieval \cite{RN30}. Sayer et al. studied the accuracy of MDT and MDB products and the fused version of MDB and MDT (presented in the 6.1 collections of MODIS (Moderate Resolution Imaging Spectroradiometer) data). They found any of these products are not entirely excellent. Therefore, the proposal to use a specific product mainly depends on the land types of the target area \cite{RN31}. Wang et al. evaluated the performance of MDB, MDT, and VDB (VIIRS Deep Blue) products. The evaluation illustrated that the VDB product had the highest accuracy in total. However, product performance varied in different land covers. For example, VDB was better than MDB in agricultural and urban areas, while MDB was more accurate in dry regions. In terms of spatial coverage, MDB and VDB products had more coverage in dry areas. In contrast, MDT had better coverage in croplands \cite{RN33}. All the mentioned studies demonstrated that improving the accuracy and spatial coverage of AOD products would affect the accuracy of PM2.5 estimation.

One way to increase the quality of various AOD products is to fuse them. 
Data fusion methods combine data from multiple sensors to improve accuracy and achieve more comprehensive information \cite{RN35} 
In this study, different fusion techniques are designed and implemented for the fusion of various AOD products from MODIS and VIIRS (Visible Infrared Imaging Radiometer Suite) observation to improve the accuracy and the spatial coverage of PM2.5 estimation ever the study area.

Previous studies have shown that fusing different AOD products can result in a product with higher accuracy and better spatial and temporal coverage \cite{RN25,RN39}. Also, the fusion of these products leads to eliminating the defects of each and increasing their potential capabilities \cite{RN29}. Xu et al. fused AODs of MISR and MODIS sensors by the maximum likelihood estimation method. The mean absolute error (MAE) of the fused product was lower than all primary AOD data. In addition, the spatial coverage of the final product was increased \cite{RN39}. Tang et al. fused AOD products of MODIS and SeaWiFS sensors using a Bayesian maximum entropy method. In the areas where both primary products had values, the accuracy of the fused product was better than the accuracy of input products. Also, the final product provided more spatial coverage. However, the results were unreliable in areas with many gaps due to insufficient observations \cite{RN25}. Han et al. fused AOD products of MODIS and CALIPO sensors in a two-step process. In the first stage, the conditions in which the AOD product of the MODIS sensor performed better than the CALIPO sensor and vice versa were determined. In the next step, the fusion of the two products was performed based on their strengths and weaknesses using conditions specified in the previous step. The accuracy of fused AOD was significantly increased compared to the original products \cite{RN41}. In another study, Wang et al. fused the AOD products of the MODIS sensor from the Terra and Aqua platforms generated by the DB algorithm using a spatial-temporal hybrid fusion model to improve the spatial coverage. The accuracy of the final product was equal to the accuracy of the input MODIS products, while the spatial coverage increased by 123\% \cite{RN42}. 

Due to the many sources of uncertainty in AOD products, estimating PM2.5 is a more complex and challenging task. It demands a more sophisticated model which can be derived from input data. For this purpose, machine learning techniques can be applied for AOD fusion to reduce uncertainties, and subsequently PM2.5 estimating. These algorithms can reduce the impact of error sources by properly adjusting weights and biases through the fusion. Compared to a wide range of classical data fusion techniques, machine learning methods automatically learn from training data and construct fusion methods with strong predictive capabilities \cite{RN35}. This way, a suitable machine learning algorithm is selected as a base model in the fusion process. In addition to selecting the base model, critical challenges should be considered in the fusion process, such as uncertainties in input data and collection methods, time differences in data retrieval, and variability in sensor technology. Thus, a suitable fusion model should comprehensively address these challenges \cite{RN36}. 

This study aims to improve the accuracy of PM2.5 estimation by eliminating or reducing the sources of uncertainties in AOD products. This will be achieved by the fusion of AOD products retrieved by different algorithms derived from various sensors. The second goal of fusion is to improve the spatial coverage of PM2.5 by reducing the number of nan-value pixels in AOD products. In previous studies, the fusion of different AOD products has generally been accomplished at the data level to produce a higher-quality AOD product. While in this study, AOD fusion is achieved at the data and decision levels with the goal of PM2.5 estimation, which was not discovered in former studies. For this purpose, an appropriate machine learning-based estimator will be employed to estimate PM2.5 concentrations through the AOD fusion at data and decision levels. In more detail, the AOD fusion at the data level will be fulfilled using the quality of the AOD retrievals delivered along with the AOD product. This idea will lead to a more accurate estimate of PM2.5 by reducing the impact of poor-quality data and involving more accurately retrieved AOD data in the fusion process.

The performance of different AOD products on PM2.5 estimates in various land types is investigated. Next, a decision level fusion based on the tuned machine learning model is devised and implemented for PM2.5 estimation from different AOD products. Specifically, the study area is Tehran, a metropolitan that suffers from severe air pollution, particularly in the cold season. While the ground air quality monitoring station partially covers the study area, they are insufficient for modeling variations of PM2.5 over the study area. Several studies have focused on AOD-PM2.5 modeling for the accurate mapping of PM2.5 over the city \cite{RN53}. In these investigations, different AOD products have been employed for PM2.5 estimation. However, no study in the literature has reported any AOD fusion strategy for better mapping of PM2.5 over Tehran. In this regard, another focus of this paper is to improve the accuracy and spatial coverage of PM2.5 maps by fusion of various AOD products.


\section{Study Area}\label{study area}
\begin{figure*}[t]
\begin{center}
\includegraphics[width=0.8\textwidth]{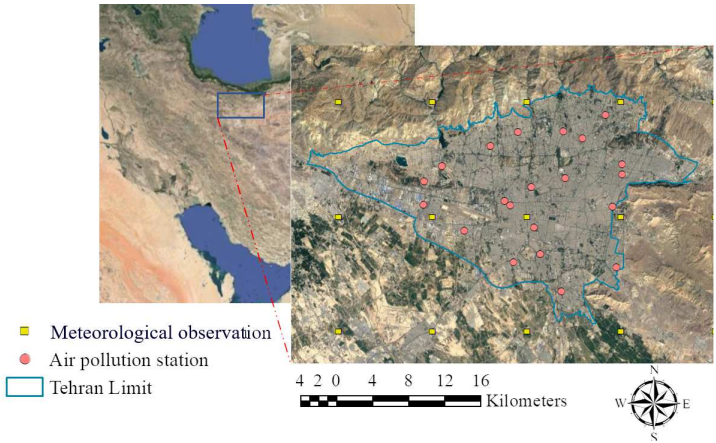}    
\caption{A display of the study area, the locations of air quality monitoring stations (pink circles), and also, the positions of meteorological observations (yellow rectangles)}  
\label{fig1}                                 
\end{center}                                 
\end{figure*}

The study area is Tehran (Shown in Figure \ref{fig1}), the capital of Iran, with a population of 13.3 million \cite{RN45}. The city is located in 35.7$ ^{\circ}  $ northern and 51.4$ ^{\circ}  $ eastern and has an area of 751 square kilometers. Regarding natural roughness, Tehran is divided into plain and foothill areas, and its current height range extends from an altitude of 900 to 1800 meters above sea level. The primary sources of air pollution in Tehran are land-use changes in recent decades, extensive industrial activities, especially in the suburbs, high population density, and an old public transportation system \cite{RN46}.

\section{Materials}\label{mat}
This study used AOD products of MODIS and VIIRS sensors, ground PM2.5 measurements, and meteorological parameters. Data were collected over seven years, from January 2013 to December 2019. The details of each data are given in the following. Also, the description, source name, and spatial resolution of parameters used to estimate PM2.5 are summarized in Table \ref{tab1}.
\begin{table*}[tb]
\centering \footnotesize \caption{Features used for PM2.5 estimation.}
\makebox[\textwidth]{\begin{tabular}{c|c|c|c}
\textbf{Data} & \textbf{Description} & \textbf{Data Source} & \textbf{Resolution} \\\hline

AOD & Aerosol Optical Depth & \thead{ MODIS (Aqua and Terra) \\ / VIIRS (Suomi NPP)} & Table 2 \\
East & \thead{East position of \\ the air pollution station (m)} & AQCC & --- \\
North & \thead{North position of \\ the air pollution station (m)} & AQCC & --- \\
DPT & dewpoint temperature (K) & ECMWF-ERA5 & 10 km \\
T & Temperature (K) & ECMWF-ERA5 & 10 km \\
Blh & Planetary boundary layer height (m) & ECMWF-ERA5 & 10 km \\
SP & Surface pressure (Pa) & ECMWF-ERA5 & 10 km \\
Lai hv & Leaf area index, high vegetation & ECMWF-ERA5 & 10 km \\
Lai lv & Leaf area index, low vegetation & ECMWF-ERA5 & 10 km \\
WS & Wind speed (m $s^{-1}$) & ECMWF-ERA5 & 10 km \\
WD & Wind direction (R) & ECMWF-ERA5 & 10 km \\
Cdir & Clear sky direct solar radiation at surface (J $m^{-2}$) & ECMWF-ERA5 & 10 km \\
Uvb & Downward UV radiation at the surface (J $m^{-2}$) & ECMWF-ERA5 & 10 km \\
RH & Relative humidity  (\%) & ECMWF-ERA5 & 10 km  \\
DOY & Day of year & - & - \\
\end{tabular}\label{tab1}}
\end{table*}
\begin{table*}[tb]
\centering \footnotesize \caption{AOD products employed in this study.}
\makebox[\textwidth]{\begin{tabular*}{\textwidth}{l|c|c|c|c|c}
\textbf{Sensor} & \textbf{Platformn} & \textbf{Resolution} & \textbf{\thead{Orbit \\ height}} & \textbf{\thead{Swath \\ width}}& \textbf{Algorithm}\\\hline

MODIS & Terra & 10km & 705km & 2330km & \thead{DB\\(Deep\_Blue\_Aerosol\_Optical\_Depth\_550\_Land)}\\MODIS & Terra & 3km & 705km & 2330km & \thead{DT\\(Image\_Optical\_Depth\_Land\_And\_Ocean)}\\MODIS & Aqua & 10km & 705km & 2330km & \thead{DB\\(Deep\_Blue\_Aerosol\_Optical\_Depth\_550\_Land)}\\MODIS & Aqua & 3km & 705km & 2330km & \thead{DT\\(Image\_Optical\_Depth\_Land\_And\_Ocean)}\\ VIIRS & SNPP & 6km & 824km & 3040km & \thead{DB\\(Aerosol\_Optical\_Thickness\_550\_Land\_Ocean)}
\\VIIRS & SNPP & 6km & 824km & 3040km & \thead{DT\\(Optical\_Depth\_Land\_And\_Ocean)}\\
\end{tabular*}\label{tab2}}
\end{table*}

\subsection{AOD Data}
In this research, AOD products of MODIS and VIIRS sensors were used. Both sensors retrieve the AOD with Dark Target (DT) and Deep Blue (DB) algorithms at 550 nm. The MODIS sensor is on the Terra and Aqua platforms, and the VIIRS sensor is on the Suomi NPP (SNPP) platform. The products of both sensors with the daily temporal resolution are available for free (\url{https://ladsweb.modaps.eosdis.nasa.gov}, \url{https://earthexplorer.usgs.gov}). Despite the similarities, these two sensors also have differences that make their AOD products different. As illustrated in Table \ref{tab2}, the platform orbit on which the VIIRS sensor is located is 824 km above the ground. In contrast, the orbit height of the MODIS sensor platforms is 705 km above the ground. Also, the swath width of the VIIRS sensor is 3040 km, while the swath width of the MODIS sensor is 2330 km. The spatial resolution of DB and DT products is 6 km for VIIRS, and 10 and 3 km for MODIS DB and DT, respectively.

All used AOD products also have a retrieval quality flag (Q). Q values vary from 0 (lowest quality) to 3 (highest quality). These values are assigned based on the number of input pixels for AOD retrieval, proximity to bright surface areas, and the amount of retrieval error. Table \ref{tab2} provides information on the AOD products used. More details are provided in the corresponding aerosol product user guides \cite{RN47}. 
\subsection{PM2.5 Measurements}
The average daily amount of PM2.5 in the study area is provided by the Tehran Air Quality Control Company (AQCC) (\url{https://www.iqair.com}). As shown in Figure \ref{fig1}, there are 21 PM2.5 measuring stations in Tehran. Some of these stations may not report PM2.5 data on some days due to technical issues. Also, the distribution of these stations is such that it covers most of the central part of Tehran, which limits the accuracy of PM2.5 estimation in areas not covered by ground measurements.
 \subsection{Meteorological Data}
Various studies have highlighted the need to use meteorological data alongside AOD data to estimate PM2.5 concentrations \cite{RN77,RN56}. In this study, the outputs of the global ECMWF-ERA5 model were used to prepare daily meteorological data. Meteorological data include dewpoint temperature, air temperature, planetary boundary layer height, surface pressure, leaf area index, clear sky direct solar radiation at the surface, wind speed, wind direction, downward UV radiation at the surface, and relative humidity. As shown in Table \ref{tab1} and Figure \ref{fig1}, these data are provided with a spatial resolution of 10 km.
\section{Experimental Setups and Methods}\label{EX}
\begin{figure*}[t]
\begin{center}
\includegraphics[width=0.8\textwidth]{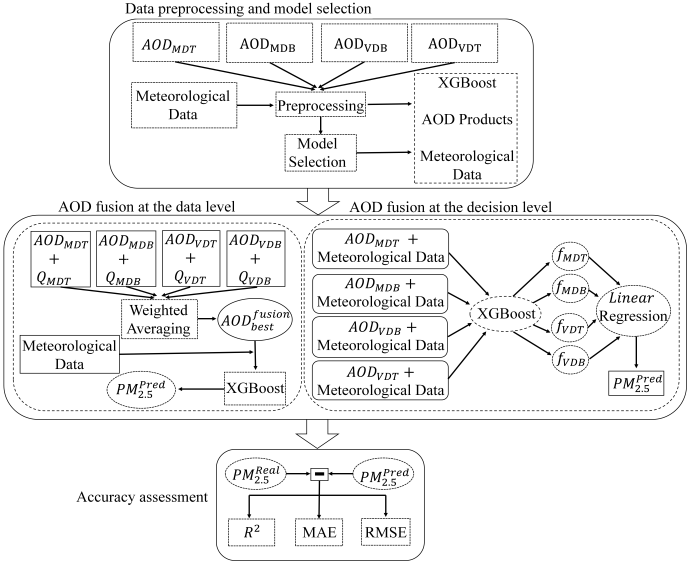}    
\caption{The framework implemented for AOD fusion at data and decision levels for PM2.5 estimation.}  
\label{fig2}                                 
\end{center}                                 
\end{figure*}

This paper investigates the effect of fusing AOD products on improving the accuracy of PM2.5 estimation by performing several experiments. Figure \ref{fig2} shows the general workflow of AOD fusion implemented in this paper. First, a preprocessing step is performed on raw AOD and meteorological data. Then, the performance of several machine learning algorithms in estimating PM2.5 using different products is evaluated. This phase aims to select the algorithm with the highest performance as a base model in the fusion of AOD products. Finally, accurate PM2.5 estimation is achieved through the devised AOD fusion at two levels of data and decision. This general structure can be deployed to fuse AOD products and estimate PM2.5 at any arbitrary location after model preparation. More details of the different stages of the devised fusion framework are explained in the following.

\subsection{AOD Preprocessing}\label{4.1}
Due to the transverse overlapping of observations taken from two adjacent orbits by satellite sensors, AOD values were averaged in overlapping areas. Then, to extract AOD in the position of each ground station, a window with a size of 3×3 pixels was used, and in each window, the mean AOD value and its standard deviation were calculated. If the standard deviation in each window was more than 0.02, the measurement was assumed unreliable and then deleted \cite{RN7}. The final AOD value in a window was determined based on the average of the reliable values (after checking the mentioned condition). In this way, we validated the measured AOD using neighboring pixels. A large difference between several neighboring pixels means the presence of destructive factors such as clouds, snow, smoke, etc. which implies the unreliability of AOD retrieval.

\begin{figure}
\begin{center}
\includegraphics[height=4cm]{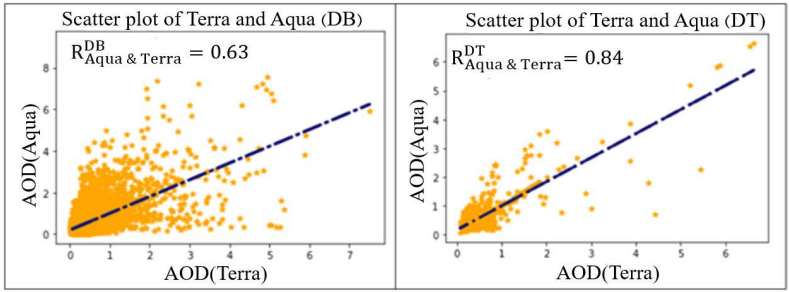}    
\caption{Correlation between Aqua and Terra AODs. Due to the high correlation between the measurements of the two sensors, a linear regression equation was used to retrieve missing AOD values for one sensor by another one.)}  
\label{fig3}                                 
\end{center}                                 
\end{figure}

Hu et al. \cite{RN19} and Lee \cite{RN49} showed that the measurements of Aqua and Terra platforms could be jointly used as daily AOD. Figure \ref{fig3} shows the scatter plots of the DB and DT products of the Aqua and Terra platforms. As shown in the figure, the products of these two platforms have a high linear correlation ($R_{Aqua-Terra}^{DB}$ =0.63 and $R_{Aqua-Terra}^{DT}$=0.84). This correlation will be used for filling in missing AOD values in one product (due to cloud, snow, etc.) while the AOD values are available in another related product. Accordingly, for points where the AOD is measured by at least one platform, it can be estimated for another platform using linear regression. Finally, the daily AOD measurement is determined by an averaging as follow:

\begin{equation}\label{eq.1}\begin{split}
AOD_{MODIS}^{DT} = \frac{(AOD_{Aqua}^{DT} + AOD_{Terra}^{DT})}{2} \\ AOD_{MODIS}^{DB} = \frac{(AOD_{Aqua}^{DB} + AOD_{Terra}^{DB})}{2},
\end{split}
\end{equation}

where, $AOD_{Aqua}^{DT} $ and $AOD_{Terra}^{DT}$ denote DT products derived from Aqua and Terra observations, respectively, while $AOD_{Aqua}^{DB}$ and $AOD_{Terra}^{DB}$ are DB products of Aqua and Terra, respectively. $AOD_{MODIS}^{DT}$ and $AOD_{MODIS}^{DB}$ are the calculated daily measurements of AODs of DT and DB products, respectively. It should be noted that this preprocessing of AOD products is only employed for data preparation through the decision level fusion. It is not applicable for data level fusion input data. More details will be clarified in section \ref{4.6.1}.

AOD is measured in a column of the atmosphere, while PM2.5 is measured near the ground surface. Thus, converting AOD as a column measurement to PM2.5 is another critical point in this process. Before converting AOD products to PM2.5, it is recommended to normalize the AOD value throughout the measuring column \cite{RN50}. Tsai et al. acquired this correction by dividing the AOD by the mixing layer height \cite{RN52}. Nabavi et al. and Bagheri used the height of the planetary boundary layer as an acceptable approximation of the height of the mixing layer in the study area \cite{RN7,RN53}. Thus, in this study, the AOD normalization will be performed by the height of the planetary boundary layer.
\subsection{PM2.5 Preprocessing}
The TEOM device determines the concentration of PM2.5 in Tehran air quality stations. This device measures the dry mass of PM2.5 by raising the ambient temperature to 50$ ^{\circ}$ C  before the measurement \cite{RN7}. Thus, the mass of measured PM2.5 is less than the true value. The reported value of PM2.5 can be modified as below \cite{RN54}:

\begin{equation}\label{eq.2}
PM_{c}=PM(1-\frac{RH}{100})^{-1} 
\end{equation}
where PM and $PM_{c}$ are measured and corrected PM2.5, respectively, and RH denotes the relative humidity expressed as a percentage.   
\subsection{Meteorological Data Preprocessing}
In addition to AOD, meteorological parameters also play an important role in estimating PM2.5 \cite{RN56,RN3}. In this study, ECMWF meteorological data were interpolated at the location of PM2.5 stations by the kriging interpolation method. The kriging interpolation parameters, such as the semi-variogram type and the kriging type, are determined by cross-validation \cite{RN73}.
\subsection{Data Analysis}
\begin{figure*}[t]
\begin{center}
\includegraphics[width=0.7\textwidth]{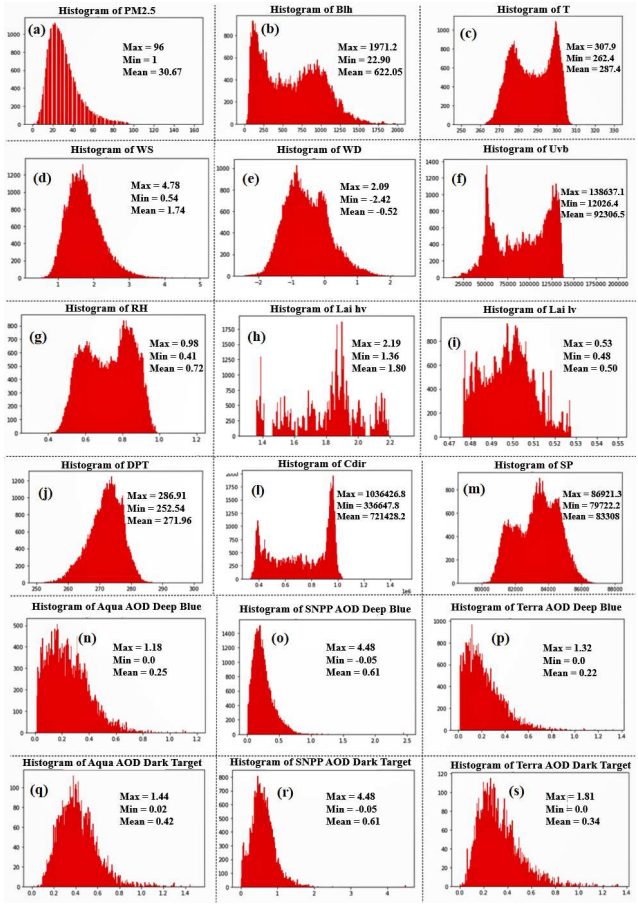}    
\caption{Histograms and descriptive statistics, including maximum (Max), minimum (Min), and mean values for all of the parameters used for PM2.5 estimation.}  
\label{fig4}                                 
\end{center}                                 
\end{figure*}

Figure \ref{fig4} shows a statistical analysis of the data used in this study. As shown in Figure \ref{fig4} (a), PM2.5 values range from 1  $\frac{\mu g}{m^{3}} $   to 96  $\frac{\mu g}{m^{3}} $ , and the graph's peak location inclines toward small data values. This means that the air quality in Tehran is mostly in moderate and unhealthy conditions for sensitive groups in terms of the air quality index (AQI). The plot of the Blh parameter (Figure \ref{fig4} (b)) also shows intensive changes from around 23 m to 1971 m. Increasing Blh increases the height of the boundary layer, decreasing the PM2.5 concentration on the ground surface. Conversely, reducing its value causes a concentration of pollutants on the ground surface. Also, temperature, an influential factor in PM2.5 transfer, varies from -10$ ^{\circ}$c to 35$ ^{\circ}$c (Figure \ref{fig4} (c)). Figure \ref{fig4} (n-s) also illustrates the distribution of retrieved AOD values from various sensors and algorithms. Generally, in different products, AOD values vary from 0 to about 5, while most measurements are between 0 and 1. The different mean values of these products indicate the difference in the accuracy of each algorithm in retrieving AOD values. Statistical parameters related to other variables are also shown in Figure \ref{fig4}.

\begin{figure*}[t]
\begin{center}
\includegraphics[width=0.8\textwidth]{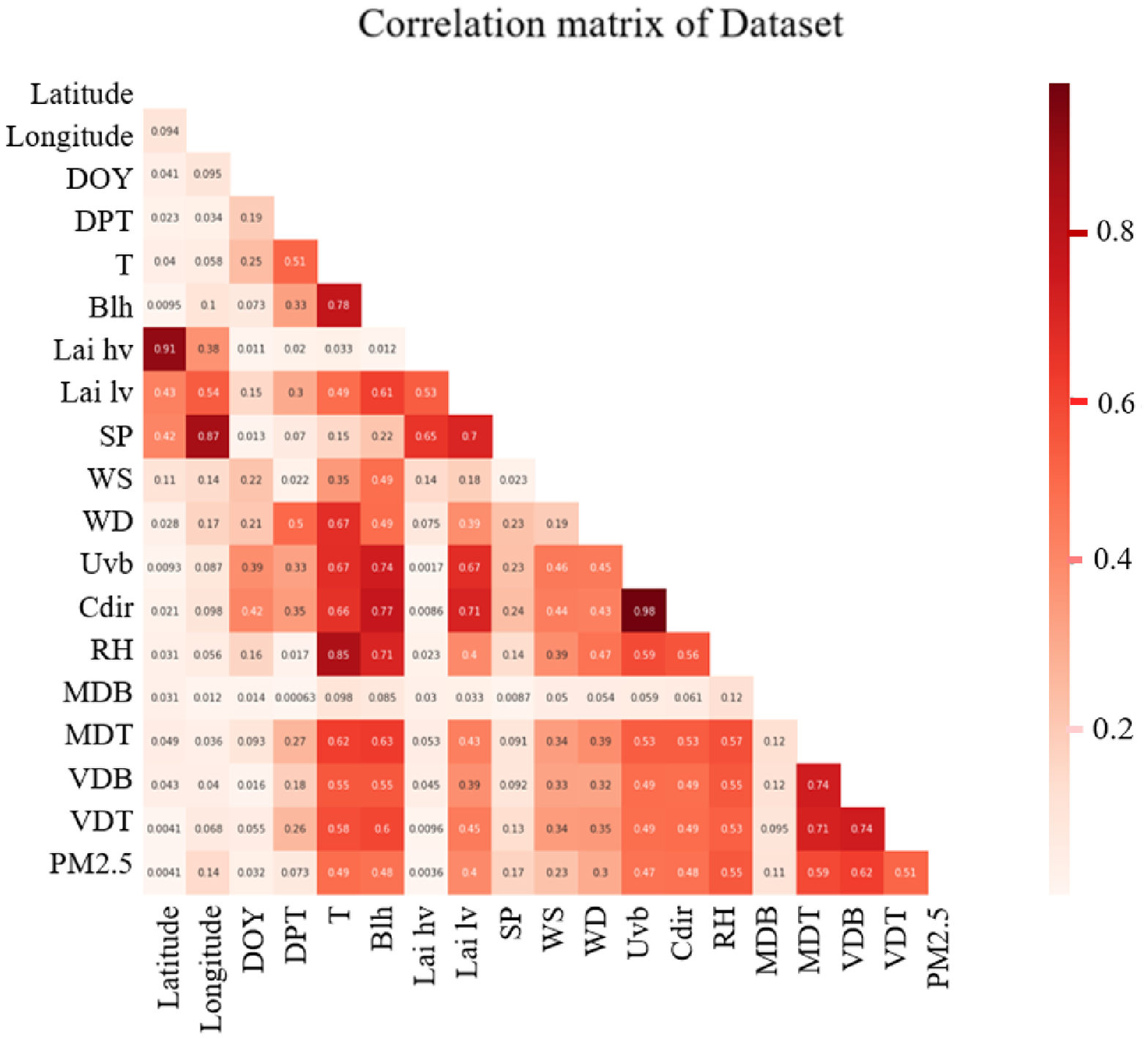}    
\caption{The correlation matrix of input parameters and PM2.5. This matrix shows the linear correlation between the variables used to estimate PM2.5.}  
\label{fig5}                                 
\end{center}                                 
\end{figure*}

Figure \ref{fig5} shows a heat map of the correlation matrix between the employed variables. In this figure, the bold red pixels indicate a high correlation between the two variables. A number within each pixel is the correlation coefficient (R) value. Some variables have a high linear correlation with PM2.5, such as VDB AOD, MDT AOD, relative humidity, VDT (VIIRS Dark Target) AOD, temperature, Blh, Cdir, Uvb, Lai-lv, and wind direction. The linear correlation of other parameters with PM2.5 is less than 0.3. However, these parameters may have a nonlinear relationship with PM2.5, which can be explored by the advanced machine learning algorithm.

\subsection{Model Selection}\label{4.5}
As mentioned in Introduction, different machine learning algorithms have been used to estimate PM2.5 from AOD data, each of which has shown different performances in various studies depending on the study area and input data. Therefore, evaluating the performance of several widely used machine learning algorithms in estimating PM2.5 using different AOD products is necessary for the first step. The goal is to find the most efficient algorithm to build a base model for fusing various AOD products.

One of the basic algorithms for estimating PM2.5 is the multivariate linear regression model \cite{RN18,RN19,RN20}. However, multivariate linear regression cannot solve complex nonlinear problems with large dimensions. More advanced algorithms, such as support vector machines (SVM), can be applied to significantly overcome issues such as complex nonlinearity and large input data dimensionality \cite{RN17}. SVM takes inseparable features from a low-dimension space to a new high-dimension space and uses a hyperplane to separate them \cite{RN61}. Support vector regressor (SVR) is a version of SVM that deals with regression problems. In addition to the SVR, decision trees have shown a high ability to solve nonlinear regression problems. Advanced versions of decision trees, such as random forest (RF), have been used in many studies to estimate PM2.5 concentrations from AOD data \cite{RN23,RN21,RN22}. RF is partially resistant to over-fitting by generating an ensemble of decision trees. In addition, RF can evaluate the importance of input variables by identifying outlier data \cite{RN66}. However, as the number of trees and the complexity of each tree rise, the training and modeling time increase. Alternatively, gradient boosting trees such as Extreme Gradient Boosting (XGBoost) algorithm is more resistant to over-fitting and use parallel processing during training, increasing processing speed and achieving higher accuracy \cite{RN67}.

Recently, deep learning algorithms have shown very successful performance in the regression and classification of various data, such as image, text, and video processing. The most common structure in solving regression problems using tabular data is the Fully Connected (FC) network. A FC network consists of several fully-connected layers, each layer being a function of $R^{n}$  to  $R^{m}$. The input of each layer is multiplied by the weights. It is then summed with a bias, and after applying a nonlinear function, the output is generated \cite{RN68}. However, deep learning algorithms generally perform weaker in the face of heterogeneous tabular data than other classical machine learning algorithms, especially decision tree-based methods. Several deep learning structures that adopt the advantageous properties of ensemble decision tree approaches, such as Neural Oblivious Decision Ensemble (NODE), have been designed to solve this issue. The NODE algorithm generalizes a set of oblivious decision trees. This structure uses both the optimization power based on gradient reduction and the power of multi-layered hierarchical learning \cite{RN69}.

\subsection{Data Level and Decision Level AOD Fusion for PM2.5 Estimation}\label{4.6}
As mentioned earlier, this paper aims to improve the PM2.5 estimation based on the fusion of AOD data. As will be shown in section 5.2, using various AOD products leads to estimating PM2.5 concentrations with different accuracies in multiple regions. These differences are induced from specific hypotheses assumed in AOD retrieval algorithms. Also, the differences in preliminary observations due to varieties in the physics of the satellite sensors can lead to AODs with different qualities \cite{RN25}. The ultimate goal in each AOD retrieval algorithm is to separate the reflection of aerosol particles from the reflection of the surface \cite{RN47}. For example, the DB product uses the blue wavelength to retrieve AOD because of the low reflection of this wavelength on bright surfaces (such as desert areas). In comparison, the DT product uses low surface reflectance in vegetated areas to retrieve AOD. Also, varying heights of the orbits of different satellites cause differences in spatial resolution and coverage of the final products \cite{RN70}. Thus, fusing different AOD products can firstly lead to a product that has better performance in estimating PM2.5 on both light and dark surfaces.

Moreover, fusion partially solves the problem of nan-value pixels in primary products caused by clouds or snow and consequently improves the spatial coverage of PM2.5 maps. Improving the spatial coverage of AOD products is essential for mapping PM2.5 variation, especially in urban areas. The following sections describe details of AOD fusion at two levels: the data level and the decision level.

\subsubsection{Fusing AOD Products at the Data Level}\label{4.6.1}

As will be shown in section \ref{R5.2}, using different AOD data leads to estimating PM2.5 with dissimilar accuracy at various air pollution measuring stations. The differences in accuracy at various points indicate that no single AOD product is ideal for evaluating high accuracy PM2.5 in all regions of the study area, so a fusion of multisource AOD products is required to achieve higher accuracy. For this purpose, we developed a data-level fusion strategy without feature extraction. The main challenge in this fusion strategy is decreasing the signal-to-noise ratio due to the use of different data and the diversity of error sources.

\begin{figure}
\begin{center}
\includegraphics[height=8cm]{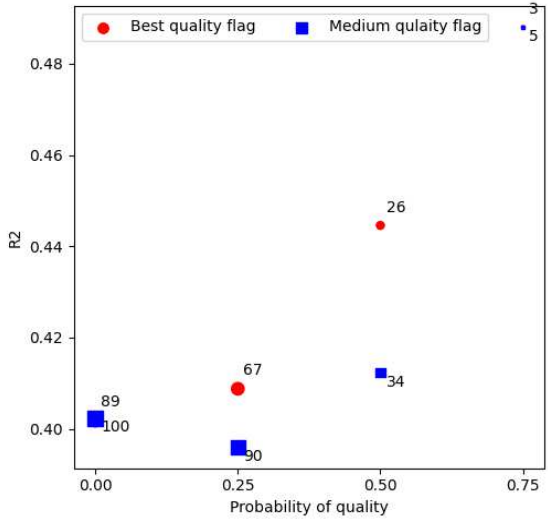}    
\caption{The influence of the probability of quality of AOD retrievals on the accuracy of PM2.5 estimation (more details can be found in \cite{RN7}.}  
\label{fig6}                                 
\end{center}                                 
\end{figure}

Previous studies have shown that filtering AOD products based on the retrieval quality (Q) of AOD improve the accuracy of PM2.5 estimation \cite{RN7,RN74}. Figure \ref{fig6} shows the results of a study on the effect of applying the MAIAC AOD retrieval quality on the PM2.5 estimation \cite{RN7}. In \cite{RN7}, AOD data were filtered based on two conditions related to their retrieval quality. The first condition includes pixels that are not cloudy and snowy (medium quality). While the second condition considers pixels with no snowy and cloudy pixels in their neighborhoods (best quality) in addition to satisfying the first condition. As shown in Figure 6, increasing the presence probability of highly qualified pixels based on the mentioned conditions, either the first or the second conditions, increases the accuracy of PM2.5 estimation according to the $R^{2}$ metric. However, filtering the data based on the aforementioned conditions reduces the number of AOD samples.

In this study, we utilized the quality of AOD retrievals to use them at the data level fusion. For this purpose, we exploited a weighted averaging method. First, we considered the probability of the presence of high-quality AOD values in the window of AOD extraction (a 3×3 window size, see section \ref{4.1} for more detail) as the weight of extracted AOD. In more detail, the weights are achieved by the ratio of the number of pixels with the best quality (Q == 3) to the total number of pixels in a 3×3 window size as follows:

\begin{equation}\label{eq.3}
    P(AOD)=  \frac{(\Sigma_{(i=1)}^{9}n)}{9}   \ \ \   n= \{ _{ 0 \ \ else}^{1 \ \      if \ \ \ Q_{i}==3}
\end{equation}

where $Q_{i}$ indicates the quality of AOD retrieval for pixel i in a 3×3 window, and P(AOD) denotes the probability of the presence of a high-quality pixel ($Q_{i}$=3) in the window. After determining the weights for AODs, the final fused AOD is calculated as eq. \ref{eq.4}:

\begin{equation}\label{eq.4}
AOD^{fused}= \frac{P(AOD_{Terra}^{DT})AOD_{Terra}^{DT}+\dots+P(AOD_{SNPP}^{DB})AOD_{SNPP}^{DB}}{P(AOD_{Terra}^{DT})+\dots+P(AOD_{SNPP}^{DB})}
\end{equation}

where the Terra, Aqua, and SNPP indices refer to the platforms of each product. The DT and DB notations also represent the AOD retrieval algorithms. A notation such as P($AOD_{Terra}^{DT}$) is the weight obtained for the DT product of the Terra platform. It should be noted that the input AOD in eq.\ref{eq.4} is the average of the AODs in a 3×3 window for each product. $AOD^{fused}$ also denotes the fused AOD product achieved from the weighted averaging.

Finally, the fused AOD and meteorological data are injected into a machine learning model such as XGBoost to estimate PM2.5.

\subsubsection{Fusing AOD Products at the Decision Level}
As explained earlier, another level of fusion is the decision level fusion. First, a base model is developed and tuned at the decision level for each AOD product to estimate PM2.5. This study selects the XGBoost algorithm as a base model to generate decisions (initial estimation of PM2.5 from each AOD product). In more detail, we import each AOD product and meteorological data into an XGBoost model tuned with appropriate hyper-parameters. The output of this process is the initial estimates of PM2.5 at each location, which include uncertainties originating from the input AOD product. To reduce these uncertainties, the estimates from the previous step, i.e., the output of XGBoost models, are given into a linear regression model as a decision maker to weight each estimation properly as below:

\begin{equation}\label{eq.5}
PM_{2.5} = A_{1}f_{XGB}^{MDB}+A_{2}f_{XGB}^{MDT}+A_{3}f_{XGB}^{VDB}+A_{4}f_{XGB}^{VDT} + B 
\end{equation}

where $A_{i}$ (i=1,2,3,4) are linear regression coefficients, B is the bias, and $f_{XGB}^{MDB}$, $f_{XGB}^{MDT}$, $f_{XGB}^{VDB}$, and $f_{XGB}^{VDT}$ are the primary estimates (decisions) obtained for MDB, MDT, VDB, and VDT products, respectively, as outputs of the tuned XGBoost models.

In this study, we considered 11 possible scenarios for the fusion of AOD products at the decision level based on the input AOD products. Table \ref{tab3} represents the compounds of AOD products used in each scenario. The × sign indicates the absence, and the $\surd$ sign indicates the presence of the product in a scenario. For example, in scenario 1, both VDB and MDB products must have valid values at a target pixel, while in scenario 2, all three products, i.e.  VDB, MDB, and MDT must be valid at that pixel. This ultimately makes the number of valid AODs for each product to be different in each scenario. Since the number of samples (valid AODs) can differ, the training and test data vary in different scenarios, subsequently. Thus for obtaining a more accurate result, the optimal hyperparameters should be explored for each scenario.
\begin{table}[tb]
\centering \footnotesize \caption{Different scenarios (input AODs) for AOD fusion at the decision level.}
\begin{tabular}{l|c|c|c|c}
\textbf{Scenario} & \textbf{MDB} & \textbf{MDT} & \textbf{VDB} & \textbf{VDT}\\\hline

1 & $\surd$ & × & $\surd$ & × \\ 2 & $\surd$ & $\surd$ & $\surd$ & × \\ 3 & $\surd$ & × & $\surd$ & $\surd$ \\ 4 & $\surd$ & $\surd$ & $\surd$ & $\surd$ \\ 5 & × & $\surd$ & × & $\surd$  \\ 6 & $\surd$ & $\surd$ & × & $\surd$  \\ 7 & × & $\surd$ & $\surd$ & $\surd$ \\ 8 & $\surd$ & $\surd$ & × & × \\ 9 & × & × & $\surd$ & $\surd$   \\ 10 & $\surd$ & × & × & $\surd$  \\ 11 & × & $\surd$  & $\surd$ & ×\\

\end{tabular}\label{tab3}
\end{table}

\section{Results}\label{res}
This section presents the results of several experiments demonstrating the effectiveness of fusing AOD products in PM2.5 estimations. For a  stricter and more detailed review of the results, in all experiments, we selected 75\% of the dataset as train data and 25\% of the dataset as test data. Hyperparameters were determined in all experiments by train data through a K-fold cross-validation (K=5) strategy. Three indicators,$R^{2}$ (coefficient of determination), RMSE (Root Mean Squared Error) and MAE (Mean Absolute Error) were used as accuracy metrics for evaluating the results.
\subsection{Results of Model Selection}
This section first evaluates the ability of machine learning algorithms introduced in section \ref{4.5} to estimate PM2.5 concentrations using different AOD products (Table \ref{tab4}). As illustrated in Table 4, the XGBoost algorithm performs best based on the $R^{2}$ and RMSE metrics for all four products. The performance of the XGBoost algorithm using the MDT product was 2\%, 0.17 $\frac{\mu g}{m^{3}} $, and 0.33 $\frac{\mu g}{m^{3}} $  better in the $R^{2}$,  RMSE, and MAE metrics than the RF algorithm. However, the accuracy of XGBoost is slightly less than the FC algorithm for this product according to the MAE metric. From the point of view of other accuracy metrics, training time, and ease of determining hyperparameters, the priority is to choose the XGBoost algorithm. When XGBoost used the MDB product to estimate PM2.5, it performed better than the RF algorithm by 2\% in $R^{2}$ and 0.07 $\frac{\mu g}{m^{3}} $ in RMSE, while in terms of the MAE metric, it was slightly worse than the RF algorithm. For the VDB product, XGBoost improved the accuracy of estimation in comparison to RF by 3\% in $R^{2}$, 0.44 $\frac{\mu g}{m^{3}} $   in RMSE, and 0.57 $\frac{\mu g}{m^{3}} $  in MAE. When the VDT product was used to estimate PM2.5, XGBoost performed as same as RF based on the $R^{2}$. However, it could outperform RF by estimating PM2.5 with the RMSE value of 0.52 $\frac{\mu g}{m^{3}} $  and the MAE value of 0.56 $\frac{\mu g}{m^{3}} $  lower. 

Another noteworthy point was the tangible superiority of XGBoost over NODE as a deep learning method. Despite the remarkable success of deep learning algorithms in various fields, these algorithms do not always perform superior to classical algorithms such as gradient-boosted decision trees in dealing with heterogeneous tabular data. Also, the training time of the XGBoost method is much less than deep learning methods \cite{RN72}.
\begin{table*}[tb]
	\centering \footnotesize
	\caption{Comparing the performance of different machine learning methods for PM2.5 estimation using AOD and meteorological data.}
	\label{Tab.model}
	\makebox[\textwidth]{\begin{tabular}{l|cccc|cccc|cccc}
		
		&  \multicolumn{4}{c|}{$R^{2}$}& \multicolumn{4}{c|}{RMSE$ \frac{\mu g}{m^{3}}$} &\multicolumn{4}{c}{MAE $\frac{\mu g}{m^{3}}$}\\
		Method & MDT  & MDB & VDT & VDB & MDT & MDB & VDT & VDB & MDT & MDB & VDT & VDB\\\hline
		
		XGBoost	& \textbf{0.70} & \textbf{0.76} &	\textbf{0.74} & \textbf{0.81}	& \textbf{10.91} &	\textbf{11.06} & \textbf{10.16} &	\textbf{10.20} &	8.41 &	8.48 &	\textbf{7.54} &	\textbf{7.71}\\  RF	& 0.68 & 	0.74 &	0.74 &	0.78 &	11.08 &	11.13 &	10.68 &	10.64 &	8.74 &	\textbf{8.44} &	8.10 &	8.28\\FC &	0.64&	0.70 &	0.67 &	0.75 &	10.92 &	12.28 &	12.02 &	11.34 & \textbf{8.38} &	8.70 &	9.44 &	8.39\\NODE &	0.50 &	0.65 &	0.54 &	0.63 &	14.31 &	14.29	& 16.23 &	13.44 &	11.61 &	10.28 &	11.71 &	10.32\\SVR &	0.54 &	0.58 &	0.51 &	0.65 &	20.86 &	18.17 & 19.39 &	18.76 &	17.54 &	14.60 &	15.33 &	14.88\\Linear &	0.50 &	0.49 &	0.50 &	0.60 &	14.36 &	24.68 &	16.04 &	18.08 &	11.36 &	11.82 &	11.83 &	13.58\\
	\end{tabular}\label{tab4}} 	
\end{table*}
\begin{table*}[tb]
\centering \footnotesize \caption{ Investigating the performance of different AOD products for PM2.5 estimation in different regions of Tehran.}
\begin{tabular}{l|c|c|c|c}
\textbf{Station Id} & \textbf{Station Name} & \textbf{\thead{Winning product \\($R^2$)}} & \textbf{\thead{Winning product \\(RMSE)}} & \textbf{\thead{Winning product\\ (MAE)}}\\\hline

1 &	Aqdasiyeh &	VDB &	MDT &	MDT\\
2 &	Rose Park &	VDB &	VDT &	MDT\\
3 &	Punak &	VDB &	MDT &	MDT\\
4 &	Piroozi &	MDB	& MDT &	MDT\\
5 &	\thead{Tarbiat Modares  \\University} &	VDB &	VDB &	VDB\\
6 &	Darrous &	VDB	& MDT &	MDT\\
7 &	Setad Bohran &	VDB &	MDT &	MDT\\
8 &	Shad Abad &	MDB &	MDB &	MDB\\
9 &	Sharif University &	VDB &	MDT &	MDT\\
10 &	Ray	 & VDB &	VDT	 & VDT\\
11 &	District2 &	MDT &	MDT &	MDB\\
12	& District4 &	MDB	 & MDB &	MDB\\
13	& District10 &	MDT &	MDB &	VDT\\
14	& District11 &	MDT	 & MDT &	MDT\\
15 &	District16 &	MDB &	VDB &	MDB\\
16 &	District19 &	MDT &	MDB	 & MDB\\
17 & 	District21 &	VDT &	MDT &	MDT\\
18 & District22 &	VDT	& VDT &	VDT\\
19	& Sadr &	MDT &	VDB &	VDB\\
20 &	Golbarg &	VDB	& MDT &	MDT\\
21	 & Masoudieh &	VDT	 & VDT &	VDT\\

\end{tabular}\label{tab5}
\end{table*}

\subsection{Result of AOD Fusion}\label{R5.2}
As mentioned in section \ref{4.6}, different AOD products give different outputs in terms of accuracy in various regions. The performance of different products in all ground stations in Tehran was studied individually using the XGBoost algorithm. Table \ref{tab5} shows the winning products through the three metrics, $R^{2}$, RMSE, and MAE. As shown, AOD products have different performances in various regions. Only, in six stations (Tarbiat Modares University, ShadAbad, District4, District11, District22, and Masoudieh), one product was better in all three metrics. Elsewhere, various products have given diverse accuracy according to different metrics. This table shows that AOD products have dissimilar accuracy in estimating PM2.5 at various locations. The diversity of surface brightness values in multiple stations, varying the number of high-quality data for each product, and the differences in the physics of sensors cause the variety in the performance of products. The chief hypothesis of this study is that the fusion of these products can lead to a more accurate PM2.5 estimation.
\subsubsection{Results of AOD Fusion at the Data Level}
AOD products were fused at the data level according to the method described in section \ref{4.6.1}. Table \ref{tab6} shows the accuracy of the fused AOD product in estimating PM2.5. Regarding the $R^{2}$ metric, the proposed product was 4\% more accurate than MDT, 3\% better than VDT, and 1\% more accurate than MDB, while it had an accuracy equal to VDB. Also, according to the RMSE metric, the results are 1.87 $\frac{\mu g}{m^{3}} $  , 1.29 $\frac{\mu g}{m^{3}} $ , 1.69 $\frac{\mu g}{m^{3}} $   , and 1.66 $\frac{\mu g}{m^{3}} $  less than MDT, VDT, MDB, and VDB products in PM2.5 estimation. The performance of the fused product in terms of MAE metric is also 1.57  $\frac{\mu g}{m^{3}} $ , 1.41  $\frac{\mu g}{m^{3}} $ , 1.43  $\frac{\mu g}{m^{3}} $ , and 1.06  $\frac{\mu g}{m^{3}} $  better than MDT, VDT, MDB, and VDB, respectively. Additionally, the spatial coverage of the fused product increased by 44\% and 9\% compared to MDT and VDT products but decreased by 8\% and 7\% compared to MDB and VDB products due to the elimination of poor-quality data during the weight generation based on the quality of AOD retrievals.

\begin{table}[tb]
\centering \footnotesize \caption{Result of AOD Fusion at data level for PM2.5 estimation.}
\begin{tabular}{c|c|c|c|c}
\textbf{Data} & \textbf{$R^{2}$} & \textbf{RMSE$\frac{\mu g}{m^{3}} $} & \textbf{MAE$\frac{\mu g}{m^{3}} $} & \textbf{\thead{Coverage\\ (percent)}}\\\hline

$MDT_{XGB}$ & 0.60 &	15.38 &	11.28 &	25\% \\
$MDB_{XGB}$ &	0.63 &	15.20 &	11.14 &	\textbf{77\%} \\ $VDT_{XGB}$ &	0.61 &	14.80 &	11.12 &	60\% \\
$VDB_{XGB}$ &	0.64 &	15.17 &	10.77 &	76\% \\
$AOD^{fused}_{XGB}$ &	\textbf{0.64} &	\textbf{13.51} &	\textbf{9.71} &	69\%\\

\end{tabular}\label{tab6}
\end{table}

\subsubsection{Results of AOD Fusion at the Decision Level}
\begin{figure*}[t]
\begin{center}
\includegraphics[width=1\textwidth]{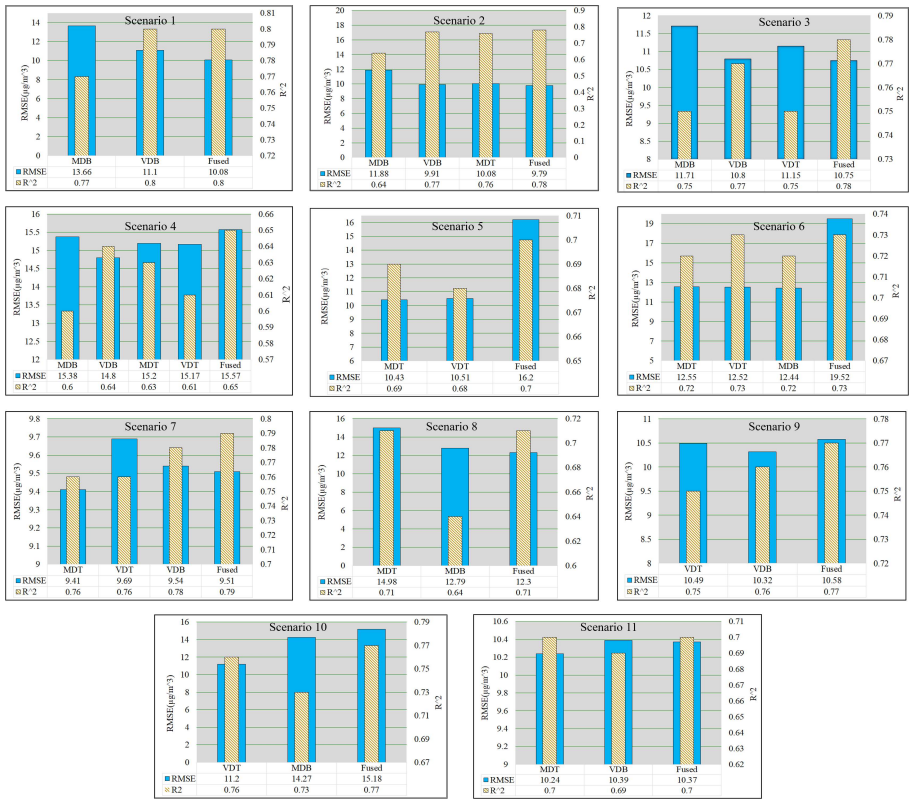}    
\caption{Results of fusing AOD data at the decision level to estimate PM2.5.}  
\label{fig7}                                 
\end{center}                                 
\end{figure*}

\begin{figure}
\begin{center}
\includegraphics[height=6cm]{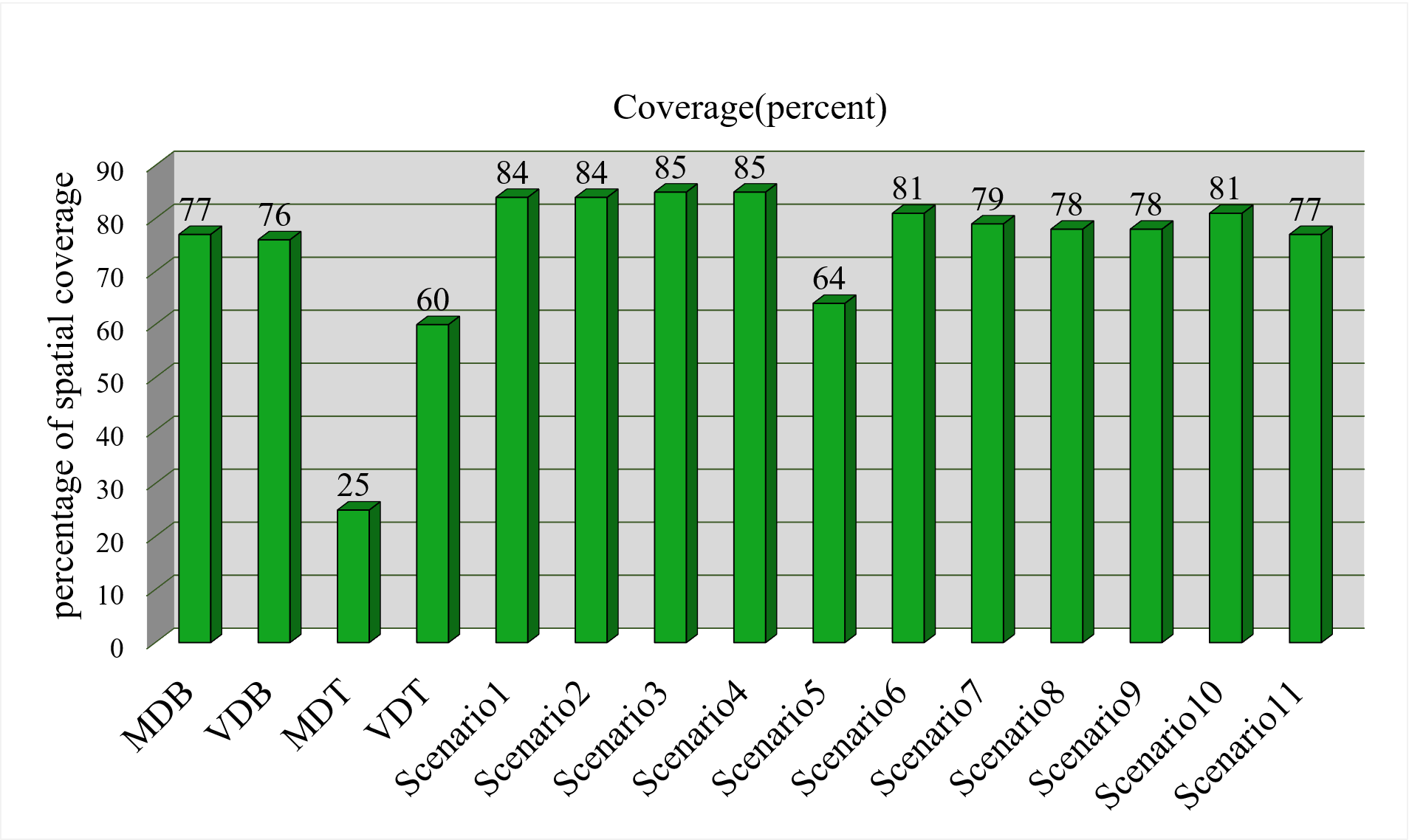}    
\caption{Improvement of spatial coverage by fusing data at the decision level. The X-axis represents the name of the scenario, and the Y-axis represents the percentage of spatial coverage.}  
\label{fig8}                                 
\end{center}                                 
\end{figure}

This study also performed AOD fusion at the decision level in eleven scenarios. Figure \ref{fig7} shows the accuracy of individual and fused products based on the RMSE  and $R^{2}$ metrics. Figure \ref{fig8} also shows the improvement of spatial coverage in each scenario. In scenario 1, the fusion approach could improve the PM2.5 estimation accuracy for the MDB and VDB products from 13.66  $\frac{\mu g}{m^{3}} $  and 11.1  $\frac{\mu g}{m^{3}} $ , respectively, to 10.08  $\frac{\mu g}{m^{3}} $ based on RMSE. In addition, $R^{2}$ metric was 3\% better than MDB and equal to VDB. Also, spatial resolution improved for the primary products from 77\% and 76\% to 84\%. Scenario 2 appends MDT products to the previous inputs. The accuracy of PM2.5 estimation with the fused product was improved compared with all input products, but the increase in accuracy was not significant compared to the previous scenario. In this scenario, spatial coverage also increased after fusion. In the third scenario, the VDT product was used instead of the MDT product. This scenario improved estimation accuracy and spatial coverage compared to individual products. This improvement in accuracy due to fusion was 0.05 $\frac{\mu g}{m^{3}} $ based on RMSE, which is not considerable compared to scenario 1. Spatial coverage increased by 1\% compared to scenario 1 (85\%). Also, the $R^{2}$ metric increased from 75\% and 77\% for individual products to 78\%. Scenario 4 utilized all four products and significantly improved spatial coverage over individual products. 
Also, the accuracy of the fused product was better in terms of the R2  than the individual products, while the accuracy was decreased based on the RMSE metric. In scenario 5, the two products, VDT and MDT, were fused. This scenario was accompanied by improving the spatial coverage and $R^{2}$ metric, but RMSE of PM2.5 was significantly degraded compared to VDT and MDT. Scenario 6 appends MDB products to the combination of scenario 5. In this scenario, fusion resulted in an estimate of PM2.5 with better spatial coverage, better $R^{2}$, and worse RMSE than the original products. Accuracy reduction based on RMSE was more significant than in scenario 5. Nevertheless, spatial coverage increased significantly from 64\% in scenario 5 to 81\% in scenario 6. In scenario 7, the VDB product was used instead of the MDB product. The fused product had better accuracy in term of $R^{2}$, and around 0.18  $\frac{\mu g}{m^{3}} $  and 0.03  $\frac{\mu g}{m^{3}} $  less RMSE,  respectively, than the VDT and VDB products, but RMSE was about 0.1  $\frac{\mu g}{m^{3}} $  higher than the MDT product. In this scenario, the spatial coverage improved significantly for VDT and MDT products to 79\%. In scenario 8, a fusion of MDT and MDB products was performed, which led to estimating PM2.5 with better accuracy and higher spatial coverage than MDB and MDT. RMSE of the fused product decreased 2.68  $\frac{\mu g}{m^{3}} $  and 0.49  $\frac{\mu g}{m^{3}} $  compared to MDT and MDB products, respectively. Also $R^{2}$ increased to 71\%. In scenario 9, when VDB and VDT products were fused, the estimated PM2.5 RMSE was higher than VDT and VDB (0.09  $\frac{\mu g}{m^{3}} $  and 0.26  $\frac{\mu g}{m^{3}} $, respectively), however  $R^{2}$ increased to 77\%. Also, the increase in spatial coverage after fusion was more significant than individual products, from 60\% for the VDT product and 76\% for the VDB product to 78\%. In scenario 10, the fusion of VDT and MDB products was accompanied by a increase in RMSE of the PM2.5 estimate, around 3.98  $\frac{\mu g}{m^{3}} $  and 0.91  $\frac{\mu g}{m^{3}} $, respectively. Spatial coverage was relatively improved, from 77\% and 60\% for MDB and VDT products, respectively, to 81\% for the fused product. In addition $R^{2}$ improved to 77\%. In scenario 11, MDT and VDB products were fused. The result was associated with a slight decrease in RMSE compared to the VDB product (0.02  $\frac{\mu g}{m^{3}} $) and a slight increase in RMSE compared to the MDT product (0.13  $\frac{\mu g}{m^{3}} $). In this scenario, the spatial coverage and $R^{2}$ were also increased.

\section{Discussion}\label{disc}
\subsection{Efficiency of AOD fusion at the data level}
Results of AOD fusion at the data level using the proposed method in this research illustrated that the output of AOD fusion is generally much superior to each input product in terms of the accuracy of PM2.5 estimation. However, Since the condition in eq.\ref{eq.3} removes poor quality data for weight generation, some available AOD data are discarded, subsequently decreasing the coverage.
\subsection{Efficiency of AOD Fusion at the Decision Level}
From the achieved results based on $R^{2}$, we can conclude that fusing these products increase the efficiency of the PM2.5 estimation in all scenarios. But this metric alone is not enough. Based on the RMSE metric, the fusion of AOD products is successful in some scenarios and unsuccessful in others. Thus, we can conclude that fusing DT products with DB products reduces the efficiency of the fusion method because of some factors. The factors are the low quality of the VDT product (based on the quality assessment file delivered along with the product (Q)), the poor spatial coverage of MDT data at brightness surfaces, and the assumptions used by the DT algorithm in the AOD retrieval phase. The fusion of DB products, especially in bright areas such as urban areas, significantly improves the accuracy and spatial coverage of PM2.5 estimation. Table \ref{tab7} presents the efficiency of the decision fusion of DB products on PM2.5 estimation. After fusion, the accuracy of PM2.5 estimation significantly increased according to the MAE metric by 3.48  $\frac{\mu g}{m^{3}} $  and 1.18  $\frac{\mu g}{m^{3}} $  compared to the MDB and VDB products.
\begin{figure}
\begin{center}
\includegraphics[height=5cm]{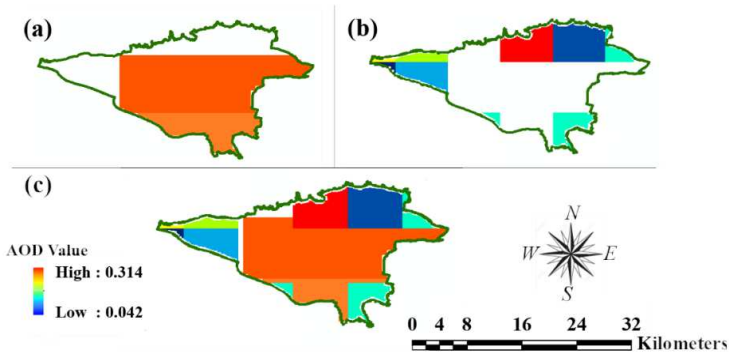}    
\caption{An example of improving the spatial coverage for AOD on the 24th of Jan. 2013 after the fusion of VDB and MDB products at the decision level. (a) MDB spatial coverage, (b) VDB spatial coverage, (c) Fused spatial coverage.}  
\label{fig9}                                 
\end{center}                                 
\end{figure}

\begin{table}[tb]
\centering \footnotesize \caption{Results of the best scenario.}
\begin{tabular}{c|c|c|c}
\textbf{Metrics} & \textbf{MDB} & \textbf{VDB} & \textbf{Fused}\\\hline

$R^{2}$	& 0.77 &	0.80 &	\textbf{0.80} \\
MAE &	10.86 &	8.56 &	\textbf{7.38}\\
RMSE &	13.66 &	11.10 &	\textbf{10.08}\\
\end{tabular}\label{tab7}
\end{table}

The DB and DT algorithms are suitable for bright and dark surface areas, as mentioned in Introduction. According to the results, due to the unsuitability of the DT algorithm for AOD retrieval in urban areas, the PM2.5 estimation RMSE does not significantly improve by fusing products in the study area. Thus, the fusion of DB-derived AODs is more applicable in the study area than other combinations of AOD products. In other words, the main reason for failing the fusion performance in scenarios involving DT products is the low accuracy of these types of AODs in urban areas, which cannot make a proper relationship with surface PM2.5 by the developed XGBoost model. However, it should be mentioned that if improving spatial coverage is a priority, all scenarios can be exploited with a slight reduction in accuracy. In other words, data fusion led to the increased spatial coverage of PM2.5 estimation. Also, the variation of PM2.5 can be visualized more accurately in urban areas by the fusion. The results indicate the effect of the selection of AOD products on the performance of the fused product in PM2.5 estimation.

\begin{figure}
\begin{center}
\includegraphics[height=6cm]{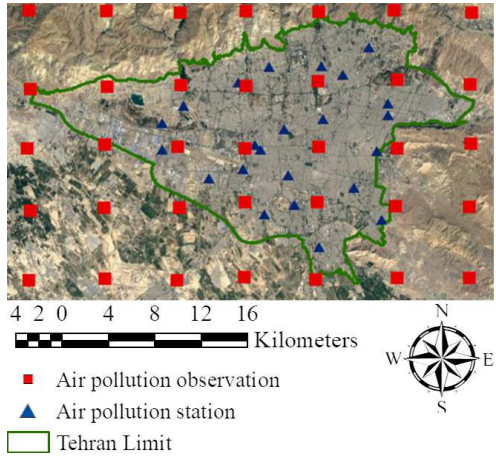}    
\caption{Distribution of PM2.5 data estimated from AOD fused product (red rectangles) and PM2.5 data measured at ground station (blue triangle) employed for PM2.5 map generation.}  
\label{fig10}                                 
\end{center}                                 
\end{figure}

Figure \ref{fig9} shows an example of fusing MDB and VDB products to improve the spatial coverage of PM2.5 estimation maps on January 24, 2013. As shown, the spatial coverage of the two products, VDB and MDB, are complementary, and their fusion has led to the estimation of PM2.5 with higher spatial coverage. Improving spatial coverage by fusion means increasing the number of PM2.5 estimates. As shown in Figure \ref{fig10}, the red rectangles are the PM2.5 estimates achieved through the fusion of VDB and MDB products. These estimates can be employed along with ground measurements of PM2.5 at the locations of air quality stations (blue triangles) to ultimately generate a more accurate PM2.5 map by an interpolation technique. In other words, improving the spatial coverage of PM2.5 estimates achieved by fusion lead to a more accurate PM2.5 map.

\begin{figure}
\begin{center}
\includegraphics[height=6cm]{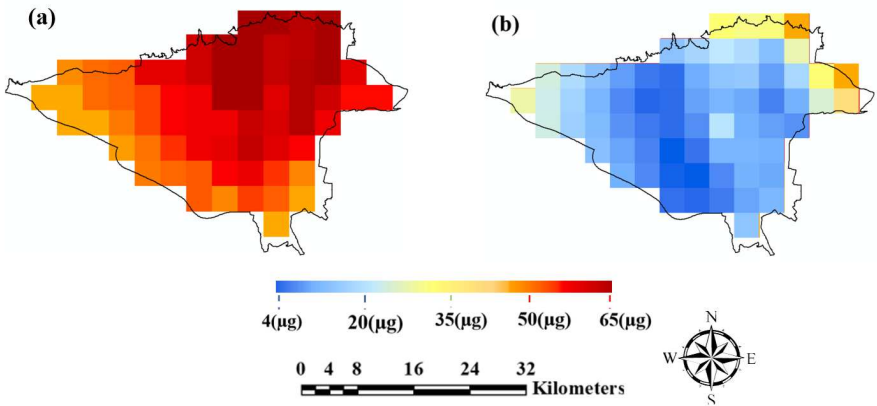}    
\caption{Daily map of PM2.5 generated by IDW interpolation of ground station measurements as well as the values estimated from fused AOD products. (a) A day with an unhealthy air quality index (29.12.2015), (b) A Day with good air quality condition (01.04.2015).}  
\label{fig11}                                 
\end{center}                                 
\end{figure}

Also, to show the performance of the proposed fusion, we plotted a PM2.5 map derived from the decision fusion of VDB and MDB products for two dates with different air quality conditions. 
Figure \ref{fig11} shows a PM2.5 map of a polluted day (December 29, 2015) and a clear day (April 1, 2015). Based on ground measurements on December 29, 2015, AQI was in unhealthy condition. Among 21 ground stations, only nine stations recorded PM2.5 concentration. To prepare this map, in addition to 9 ground points, 23 quasi-stations from MDB and VDB AOD data and meteorological information were used in this study. The values of PM2.5 in these stations were estimated by the proposed decision fusion strategy. According to Figure \ref{fig11}-a, in the central, northern, and northeastern regions of the study area, the concentration of PM2.5 particles has been estimated to be between 50  $\frac{\mu g}{m^{3}} $   and 65  $\frac{\mu g}{m^{3}} $  , which lies in the unhealthy category according to AQI categories. Based on PM2.5 measured by ground stations on April 1, 2015, the air quality index (AQI) is in the clean range. On this day, 17 stations out of 21 ground stations recorded PM2.5 concentrations. In addition to 17 ground measurements, eight quasi-stations were applied to generate the PM2.5 map this day. According to Figure \ref{fig11}-b, the concentration of PM2.5 in the central areas covered by ground stations has been estimated to be between 4 to 10, which is consistent with the range of AQI announced by AQCC.

This study concluded that the fusion of various DB-derived AOD products could improve the accuracy and the spatial coverage of PM2.5 estimation rather than individual products, especially in urban areas.
\section{Conclusion}\label{con}
In this study, we performed several experiments to investigate the effect of fusing AOD products on improving the accuracy and the spatial coverage of PM2.5 estimation. First, the performance of various machine learning algorithms in estimating PM2.5 was evaluated. The results showed the superiority of the XGBoost algorithm over the study area. Consequently, the XGBoost algorithm was selected as the base model for PM2.5 estimation. In the following, through the AOD fusion, the accuracy of different AOD products in estimating PM2.5 concentration in various regions was investigated. The results revealed the necessity of fusing AOD products for accurate PM2.5 estimation. Thus, the fusion of AOD products was performed at two levels of data and decision. The fusion at the data level was performed based on the weighted averaging using the weights obtained from the quality of AOD retrieval. At the data level fusion, the accuracy of PM2.5 was improved compared to the original products. Still, the spatial coverage did not increase considerably due to eliminating poor-quality data during the weight generation.

Next, we fused various products at the decision level in several scenarios. Fusion at the decision level led to improving spatial coverage in all scenarios. However, in terms of accuracy improvement for the aim of PM2.5 estimation for urban areas, the best result was obtained by fusing VDB and MDB products.

\section*{Acknowledgments}
The author wishes to express his gratitude to everyone who contributed to this study, particularly Tehran's AQCC for the ground PM2.5 measurements, NASA for the MAIAC data, and ECMWF for the meteorological data.

\section*{Declarations}

\subsection*{Ethical Approval}

All authors have read, understood, and have complied as applicable with the statement on "Ethical responsibilities of Authors" as found in the Instructions for Authors and are aware that with minor exceptions, no changes can be made to authorship once the paper is submitted.
 
\subsection*{Conflict of Interest} 
The authors declare that he has no conflict of interest.
 
\subsection*{Authors' contributions}
\begin{itemize}

\item Ali Mirzaei: Methodology, Investigation, Data curation, Software, Validation, Writing- Original draft preparation, Visualization.

\item Hossein Bagheri: Conceptualization, Methodology, Software, Investigation, Supervision, Validation, Writing- Reviewing and Editing.

\item Mehran Sattari: Conceptualization, Writing- Reviewing and Editing.
\end{itemize}

\subsection*{Funding}
This research received no specific grant from any funding agency in the public, commercial, or not-for-profit sectors.

\subsection*{Availability of data and materials }
The data that support the findings of this study are available on request from the corresponding author.



\end{document}